\documentclass[sigtbd]{sigtbd17-style_nomargin}
\usepackage{animate}
\usepackage{courier}            
\usepackage{graphicx}
\usepackage{amsmath}
\usepackage{amssymb}
\usepackage[scaled]{helvet} 
\usepackage{url}                  
\usepackage{xurl}
\usepackage{listings}          
\usepackage{enumitem}      
\usepackage{lipsum}
\usepackage{scrextend}
\makeatletter
\def\@copyrightspace{\relax}
\makeatother
\begin{document}
\begin{figure}[ht]
\begin{center}
\vspace{-7em}
\animategraphics[controls,loop,autoplay, width=\paperwidth]{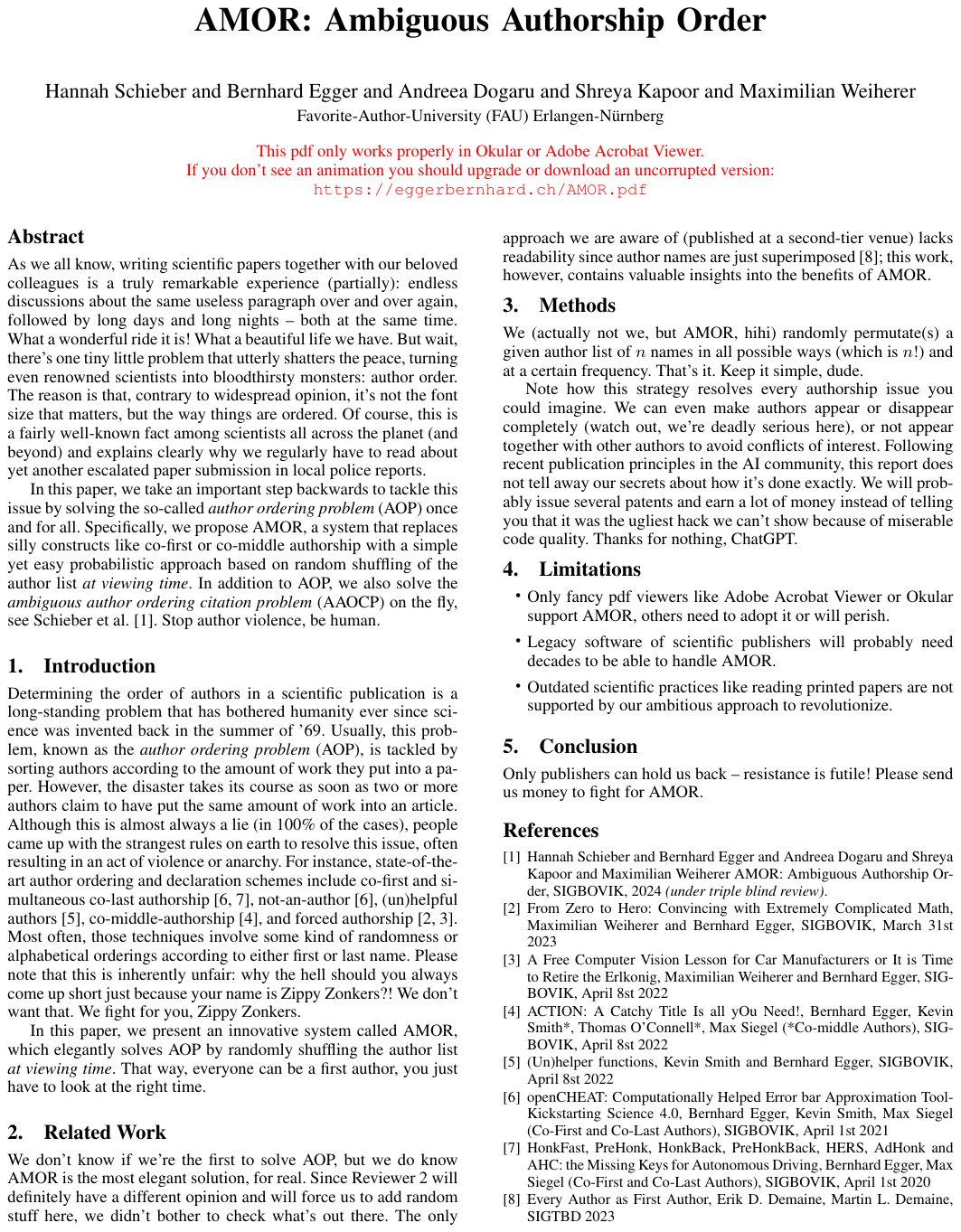}{./}{1}{120}
\end{center}
\end{figure}
\clearpage
\newpage
\pagebreak

\begin{addmargin}[7em]{0em}

The following pages are just added for parsing
\begin{abstract}
As we all know, writing scientific papers together with our beloved colleagues is a truly remarkable experience (partially): endless discussions about the same useless paragraph over and over again, followed by long days and long nights -- both at the same time. 
What a wonderful ride it is!
What a beautiful life we have.
But wait, there's one tiny little problem that utterly shatters the peace, turning even renowned scientists into bloodthirsty monsters: author order.
The reason is that, contrary to widespread opinion, it's not the font size that matters, but the way things are ordered.
Of course, this is a fairly well-known fact among scientists all across the planet (and beyond) and explains clearly why we regularly have to read about yet another escalated paper submission in local police reports.
 
In this paper, we take an important step backwards to tackle this issue by solving the so-called \textit{author ordering problem} (AOP) once and for all.
Specifically, we propose AMOR, a system that replaces silly constructs like co-first or co-middle authorship with a simple yet easy probabilistic approach based on random shuffling of the author list \textit{at viewing time}.
In addition to AOP, we also solve the \textit{ambiguous author ordering citation problem} (AAOCP) on the fly, see Weiherer et~al. \cite{aa}.
Stop author violence, be human.
\end{abstract}

\section{Introduction}
Determining the order of authors in a scientific publication is a long-standing problem that has bothered humanity ever since science was invented back in the summer of '69.
Usually, this problem, known as the \textit{author ordering problem} (AOP), is tackled by sorting authors according to the amount of work they put into a paper.
However, the disaster takes its course as soon as two or more authors claim to have put the same amount of work into an article.
Although this is almost always a lie (in 100\% of the cases), people came up with the strangest rules on earth to resolve this issue, often resulting in an act of violence or anarchy.
For instance, state-of-the-art author ordering and declaration schemes include co-first and simultaneous co-last authorship~\cite{f,e}, not-an-author~\cite{e}, (un)helpful authors~\cite{d}, co-middle-authorship~\cite{c}, and forced authorship~\cite{a,b}.
Most often, those techniques involve some kind of randomness or alphabetical orderings according to either first or last name.
Please note that this is inherently unfair: why the hell should you always come up short just because your name is Zippy Zonkers?!
We don't want that. We fight for you, Zippy Zonkers.

In this paper, we present an innovative system called AMOR, which elegantly solves AOP by randomly shuffling the author list \textit{at viewing time}. 
That way, everyone can be a first author, you just have to look at the right time.

\section{Related Work}
We don't know if we're the first to solve AOP, but we do know AMOR is the most elegant solution, for real. Since Reviewer 2 will definitely have a different opinion and will force us to add random stuff here, we didn't bother to check what's out there. The only approach we are aware of (published at a second-tier venue) lacks readability since author names are just superimposed~\cite{every}; this work, however, contains valuable insights into the benefits of AMOR.
\vspace{-0.5em}
\section{Methods}
We (actually not we, but AMOR, hihi) randomly permutate(s) a given author list of $n$ names in all possible ways (which is $n$!) and at a certain frequency.
That's it. Keep it simple, dude.

Note how this strategy resolves every authorship issue you could imagine. We can even make authors appear or disappear completely (watch out, we're deadly serious here), or not appear together with other authors to avoid conflicts of interest. Following recent publication principles in the AI community, this report does not tell away our secrets about how it's done exactly. We will probably issue several patents and earn a lot of money instead of telling you that it was the ugliest hack we can't show because of miserable code quality. Thanks for nothing, ChatGPT.
\vspace{-0.5em}
\section{Limitations}
\begin{itemize}
    \item Only fancy pdf viewers like Adobe Acrobat Viewer or Okular support AMOR, others need to adopt it or will perish.
    \item Legacy software of scientific publishers will probably need decades to be able to handle AMOR.
    \item Outdated scientific practices like reading printed papers are not supported by our ambitious approach to revolutionize.
\end{itemize}
\vspace{-0.5em}
\section{Conclusion}
Only publishers can hold us back -- resistance is futile!
Please send us money to fight for AMOR.

\end{addmargin}
\end{document}